\documentclass[letterpaper, 10 pt, conference]{ieeeconf}  

\IEEEoverridecommandlockouts                              

\usepackage{graphics}
\usepackage{epsfig}
\usepackage{amsmath,amssymb,amsfonts}
\usepackage{algorithm}
\usepackage{algorithmicx}
\usepackage{algpseudocode}
\usepackage{color}
\usepackage{url}
\usepackage{hyperref}

\newcommand{\norm}[1]{\left\lVert#1\right\rVert}
\usepackage{subfig} 
\usepackage{multicol}
\usepackage{multirow}
\usepackage{hhline}
\usepackage{physics}
\usepackage{textcomp}

\title{\LARGE \bf
Exploring Imitation Learning for Autonomous Driving with Feedback Synthesizer and Differentiable Rasterization
}

\author{Jinyun Zhou, Rui Wang, Xu Liu, Yifei Jiang, Shu Jiang, Jiaming Tao, Jinghao Miao, Shiyu Song$^{1}$
\thanks{The authors are with Baidu Autonomous Driving Technology Department, \texttt{\{jinyunzhou, ruiwang, liuxu21, jiangyifei, shujiang, taojiaming, miaojinghao, songshiyu\}@baidu.com.}}%
\thanks{$^{1}$Author to whom correspondence should be addressed, E-mail: {\tt\small \href{mailto:songshiyu@baidu.com}{songshiyu@baidu.com}}}%
}

\begin{document}

\maketitle
\thispagestyle{empty}
\pagestyle{empty}

\begin{abstract}
We present a learning-based planner that aims to robustly drive a vehicle by mimicking human drivers' driving behavior. We leverage a mid-to-mid approach that allows us to manipulate the input to our imitation learning network freely. With that in mind, we propose a novel feedback synthesizer for data augmentation. It allows our agent to gain more driving experience in various previously unseen environments that are likely to encounter, thus improving overall performance. This is in contrast to prior works that rely purely on random synthesizers. Furthermore, rather than completely commit to imitating, we introduce task losses that penalize undesirable behaviors, such as collision, off-road, and so on. Unlike prior works, this is done by introducing a differentiable vehicle rasterizer that directly converts the waypoints output by the network into images. This effectively avoids the usage of heavyweight ConvLSTM networks, therefore, yields a faster model inference time. About the network architecture, we exploit an attention mechanism that allows the network to reason critical objects in the scene and produce better interpretable attention heatmaps. To further enhance the safety and robustness of the network, we add an optional optimization-based post-processing planner improving the driving comfort. We comprehensively validate our method's effectiveness in different scenarios that are specifically created for evaluating self-driving vehicles. Results demonstrate that our learning-based planner achieves high intelligence and can handle complex situations. Detailed ablation and visualization analysis are included to further demonstrate each of our proposed modules' effectiveness in our method.
\end{abstract}
\section{introduction}
In recent years, autonomous driving technologies have seen dramatic advances in self-driving vehicles' commercial and experimental operations.
Due to complex road topologies across cities and dynamic interaction among vehicles or pedestrians, traditional planning approaches that involve heavy manual tuning are believed to be not cost-effective or scalable \cite{sadat2019jointly}.
An alternative is imitation learning, a data-driven approach that mimics expert agents' driving behavior and leverages recent advances in supervised learning.

\begin{figure}[!ht]
	\centering
	\includegraphics[width=\linewidth]{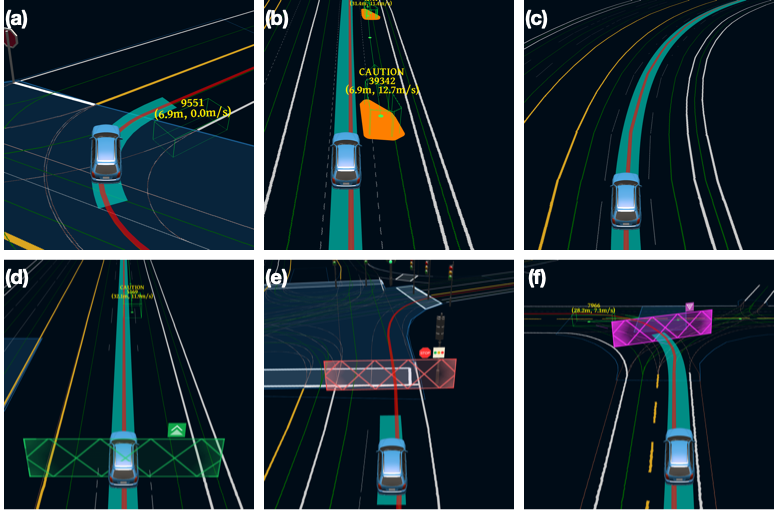}
	\caption{\small Demonstration of the test scenarios that our learning-based planner (M2) successfully passed: (a) passing a static obstacle; (b) overtaking a dynamic obstacle; (c) cruising on a curved lane; (d) following a leading vehicle; (e) stopping for a red traffic light; (f) yielding to a vehicle.
	}
	\label{fig:intro}
	\vspace{-0.4cm}
\end{figure}

Imitation learning suffers from a long bothering issue, distributional shift \cite{ross2011reduction,prakash2020exploring}.
Variation in input state distribution between training and testing sets causes accumulated errors in the learned policy output, which further leads the agent to an unseen environment where the system fails ultimately.
The key solution to this compounding problem is to increase the demonstrated data as much as possible.
Inspired by \cite{blukis2018following,bansal2019chauffeurnet,buhler2020driving,chen2020learning}, our work adopts a mid-to-mid approach where our system's input is constructed by building a top-down image representation of the environment that incorporates both static and dynamic information from our HD Map and perception system.
This gives us large freedom to manipulate the input to an imitation learning network \cite{bansal2019chauffeurnet}.
Therefore, it helps us handily increase the demonstrated data using appropriate data augmentation techniques and neatly avoid technical difficulties in building a high fidelity raw sensor data simulator.
Following the philosophy of DAgger \cite{ross2011reduction}, we introduce a feedback synthesizer that generates and perturbs on-policy data based on the current policy.
Then we train the next policy on the aggregate of collected datasets.
The feedback synthesizer addresses the distributional shift issue, thus improving the overall performance shown in Section~\ref{section:exp}.


Furthermore, pure imitation learning does not explicitly specify important goals and constraints involved in the task, which yields undesirable behaviors inevitably.
We introduce task losses to help regulate these unwanted behaviors, yielding better causal inference and scene compliance.
Inspired by \cite{wang2020improving}, The task losses are implemented by directly projecting the output trajectories into top-down images using a differentiable vehicle rasterizer.
They effectively penalize behaviors, such as obstacle collision, traffic rule violation, and so on, by building losses between rasterized images and specific object masks.
Inspired by recent works \cite{cui2020deep}, our output trajectories are produced by a trajectory decoding module that includes an LSTM \cite{hochreiter1997long} and a kinematic layer that assures our output trajectories' feasibility.
On the whole, these designs help us avoid using the heavy AgentRNN network in Chauffeurnet \cite{bansal2019chauffeurnet}, which functions similarly to a Convolutional-LSTM network \cite{shi2015_convolutional}.
It is demonstrated that our proposed method achieves a much shorter network inference time comparing to \cite{bansal2019chauffeurnet}, as shown in Section \ref{subsec:run_time}.
Moreover, to further improve the performance, similar to recent works \cite{hecker2020learning, kim2020attentional}, we introduce a spatial attention module in our network design.
The heatmap estimated by the attention module provides us a tool for visualizing and reasoning critical objects in the environment and reveals a causal relationship between them and the ego vehicle, as shown in Figure~\ref{figure:attention}.

At last, although a standard vehicle controller can directly use the trajectory waypoints generated by our network to drive the vehicle following them, we propose to add an optional post-processing planner as a gatekeeper which manages to interpret them as high-level decision guidance and composes a new trajectory that offers better comfort.
This architecture can be extended in the future, addressing the paramount safety concern caused by the fact that a neural network is typically considered a black box and hard to interpret and analyze.


Our models are trained with 400 hours of human driving data.
We evaluated our system using 70 autonomous driving test scenarios (ADS) that are specifically created for evaluating the fundamental driving capabilities of a self-driving vehicle.
We show that our learning-based planner (M2) trained via imitation learning achieves 70.0\% ADS pass rate and can intelligently handle different challenging scenarios, including overtaking a dynamic vehicle, stopping for a red traffic light, and so on, as shown in Figure~\ref{fig:intro}.
\section{related works} 
\subsubsection{Imitation Learning}
Imitation Learning for motion planning was first introduced in the pioneering work \cite{pomerleau1988alvinn} where it directly learns a policy that maps sensor data to steering angle and acceleration. In recent years, there is an extensive literature \cite{bojarski2016end, hecker2018end, codevilla2018end, bewley2019learning, codevilla2019exploring, prakash2020exploring,hecker2020learning} that follows this end-to-end philosophy. Alternatively, our work adopts a mid-to-mid approach \cite{bansal2019chauffeurnet, buhler2020driving} which allows us to augment data handily and go beyond pure imitation by having task-specific losses.

\begin{figure}
\begin{minipage}{0.16\textwidth}
 \centering
 \subfloat[Rasterized Scene \label{figure:input_in_rgb}]{\includegraphics[width=0.9\linewidth]{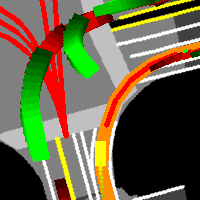}}
 \end{minipage}\hfill
\begin{minipage}{0.16\textwidth}
 \centering
 \subfloat[Agent Box \label{figure:agent_current_box}]{\includegraphics[width=0.9\linewidth]{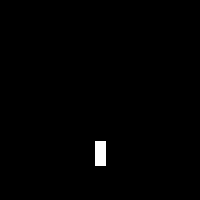}}
\end{minipage}\hfill
\begin{minipage}{0.16\textwidth}
 \centering
 \subfloat[Past Agent Poses \label{figure:agent_trajectory_history}]{\includegraphics[width=0.9\linewidth]{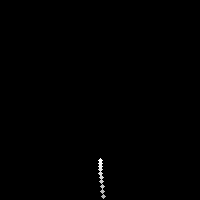}}
 \end{minipage}\hfill
\begin{minipage}{0.16\textwidth}
 \centering
 \subfloat[Obs Prediction\label{figure:obstacles_prediction}]{\includegraphics[width=0.9\linewidth]{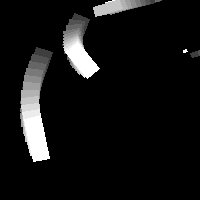}}
\end{minipage}\hfill
\begin{minipage}{0.16\textwidth}
 \centering
 \subfloat[Obs History\label{figure:obstacles_history}]{\includegraphics[width=0.9\linewidth]{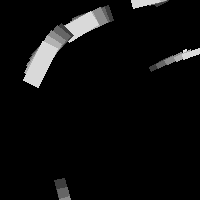}}
 \end{minipage}\hfill
\begin{minipage}{0.16\textwidth}
 \centering
 \subfloat[HD Map \label{figure:local_road_map}]{\includegraphics[width=0.9\linewidth]{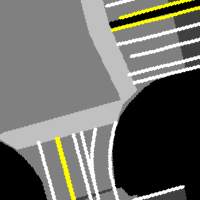}}
 \end{minipage}\hfill
\begin{minipage}{0.16\textwidth}
 \centering
 \subfloat[Routing \label{figure:local_routing}]{\includegraphics[width=0.9\linewidth]{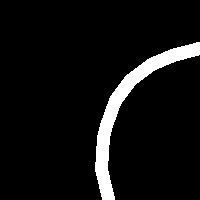}}
 \end{minipage}\hfill
\begin{minipage}{0.16\textwidth}
 \centering
 \subfloat[Speed Limit \label{figure:local_speed_limit}]{\includegraphics[width=0.9\linewidth]{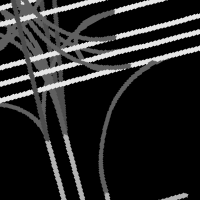}}
 \end{minipage}\hfill
\begin{minipage}{0.16\textwidth}
 \centering
 \subfloat[Traffic Lights \label{figure:traffic_light_status}]{\includegraphics[width=0.9\linewidth]{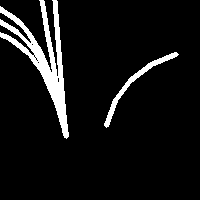}}
 \end{minipage}\hfill
 \caption{\small Scene rasterization in BEV as multi-channel image inputs for a scene shown in (a): (b) agent rasterized as a box; (c) past agent trajectory rasterized as a sequence of points with fading brightness (The oldest points are darkest); (d) prediction of obstacles rasterized as a sequence of boxes with fading brightness (The furthest boxes are brightest); (e) history of obstacles (The oldest boxes are darkest); (f) a color (3-channel) image rendered with surrounding road structures including lanes, intersections, crosswalks, etc; (g) the intended route rasterized in a constant white color; (h) lanes colored in proportion to known speed limit values; (i) traffic lights affected lanes colored in different grey levels that are corresponding to different states of traffic lights.}
 \label{figure:model_inputs_representation}
 \vspace{-0.4cm}
\end{figure}

\subsubsection{Loss and Differentiable Rasterization}
Imitation learning for motion planning typically applies a loss between inferred and ground truth trajectories~\cite{codevilla2018end,codevilla2019exploring}.
Therefore, the ideas of avoiding collisions or off-road situations are implicit and don't generalize well.
Some prior works~\cite{bhardwaj2020differentiable,bansal2019chauffeurnet} propose to introduce task-specific losses penalizing these unwanted behaviors.
With that in mind, we introduce task losses in our work and achieve real-time inference based on a differentiable rasterizer.
Wang et al.~\cite{wang2020improving} leverage a differentiable rasterizer, and it allows gradients to flow from a discriminator to a generator, enhancing a trajectory prediction network powered by GANs \cite{goodfellow2014generative}.
Different from us, concurrent work~\cite{cui2020ellipse} tackles a trajectory prediction task for better scene-compliant using similar rasterizers.
General-purpose differentiable mesh renderers~\cite{kato2018neural,liu2019soft} have also been employed to solve other computer vision tasks.

\subsubsection{Attention}
The attention mechanism has been used successfully in both computer vision~\cite{wang2017residual,hu2018squeeze,lu2019deepvcp} or natural language processing~\cite{vaswani2017attention} tasks.
Some works~\cite{kim2017interpretable,kim2019grounding,zhou2020da4ad} employ the attention mechanism to improve the model's interpretability by providing spatial attention heatmaps highlighting image areas that the network attends to in their tasks.
In this work, we introduce the Atrous Spatial Attentional Bottleneck from \cite{kim2020attentional}, providing easily interpretable attention heatmaps while also enhancing the network's performance.

\begin{figure*}
	\centering
	\includegraphics[width=\linewidth]{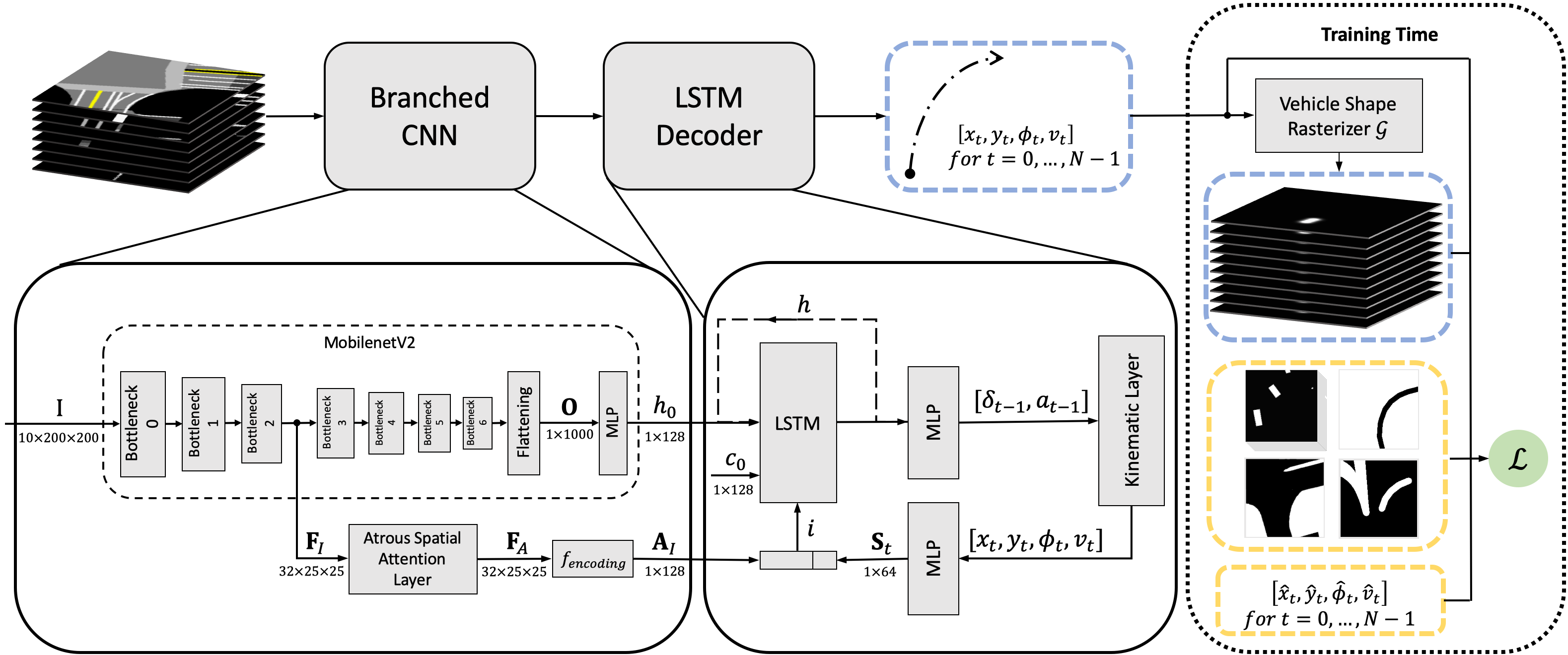}
	\caption{\small The illustration of the network structure of the three main modules: (a) a CNN backbone with a branched spatial attention structure; (b) an LSTM trajectory decoder with a kinematical layer; (c) a vehicle shape differentiable rasterizer.
	}
	\label{fig:network}
	\vspace{-0.4cm}
\end{figure*}

\subsubsection{Data Augmentation}
DAgger \cite{ross2011reduction} and its variants \cite{zhang2017query,blukis2018following,prakash2020exploring} propose to address the distributional shift issue by having more data with states that the agent is likely to encounter.
In particular, they sample new states at each iteration based on the actions inferred from the current policy, let expert agents demonstrate the actions they would take under these new states, and train the next policy on the aggregate of collected datasets.
They have been explored in the context of autonomous driving in prior works \cite{zhang2017query,pan2018agile,chen2020learning}.
ChauffeurNet \cite{bansal2019chauffeurnet} introduces a random synthesizer that augments the demonstration data by synthesizing perturbations to the trajectories.
In this work, we explore both ideas and propose a feedback synthesizer improving the overall performance.

\section{Method}
\subsection{Model Architecture}
\label{section:model}
\subsubsection{Model Input}
We form a bird`s eye view (BEV) representation with multiple channels by scene rasterization, similar to those used in both earlier planning \cite{bansal2019chauffeurnet,buhler2020driving,hecker2020learning} or prediction \cite{casas2018intentnet,chai20a,Djuric_2020_WACV} works.
The image size is \textit{W}~\texttimes~\textit{H} with $\rho$ meters per pixel in resolution.
All channels are ego-centered, which means an agent vehicle positioned at $\mathbf{x}_0\!=\![x_0, y_0]^T$ is always centered at image coordinate $\mathbf{p}_0\!=\![i_0, j_0]^T$ within the image and heading upward.
The input $\mathcal{I}$ consists of several different channels, as shown in Figure \ref{figure:model_inputs_representation}.
Both prediction or history of our ego-vehicle or other obstacles is sampled with a fixed time interval of $\Delta t$.

Besides the multi-channel images, vehicle speed $v_0$ is also an input, incorporated later by the kinematic motion model into the network (detailed in Section~\ref{subsection_model_architecture}).

\subsubsection{Model Design} \label{subsection_model_architecture}
The model consists of three parts, a CNN backbone with a branch of spatial attention module that is stemmed from intermediate features of the CNN, an LSTM decoder taking the feature embeddings from the CNN as inputs and outputting planning trajectories in the time domain, a differentiable rasterizer module that is appended to the LSTM decoder during training rasterizing output trajectories into images and finally feeding them to our loss functions. The model architecture is shown in Figure~\ref{fig:network}.

For the CNN backbone, we use MobileNetV2~\cite{sandler2018mobilenetv2} for its well-balanced accuracy and inference speed.
Then, its output features $\mathbf{F}_h$ are passed through a multilayer perceptron (MLP) layer, producing flattened feature output $\mathbf{h}_0$, which is the initial hidden state of our LSTM decoder.
Also, to lighten the computation workload, intermediate features $\mathbf{F}_{\mathcal{I}}$ from the backbone instead of raw image channels are fed to our spatial attention module.
Our attention implementation adopts the Atrous Spatial Attentional Bottleneck from~\cite{kim2020attentional}, which provides better interpretable attention maps resulting in attentive features $\mathbf{F}_{\mathcal{A}}$.
Following an encoding block $f_{encoding}$ consisting of a sequence of 3~\texttimes~3 convolutions and average poolings, a vector $\mathbf{A}_{\mathcal{I}}$ is output by the spatial attention module.
It constructs a part of the input $\mathbf{i}$ of the succeeding LSTM decoder.
The cell state $c_0$ is initialized by the Glorot initialization~\cite{glorot2010understanding}.
Then, our LSTM decoder takes these inputs and generates a sequence of future vehicle steering angle and acceleration values $(\delta_{t-1}, a_{t-1})$, where $t=0,\dots,N-1$.
Similar to prior works in the fields of vehicle control~\cite{kong2015kinematic} or trajectory prediction~\cite{cui2020deep}, we add a kinematic model that is differentiable and acts as a network layer in the LSTM.
Given the latest states $(x_{t-1}, y_{t-1}, \phi_{t-1}, v_{t-1})$, it converts $(\delta_{t-1}, a_{t-1})$ to corresponding vehicle horizontal coordinates, heading angles, and velocities $(x_t, y_t, \phi_t, v_t)$, yielding kinematically feasible planning trajectories:
\begin{equation}
\label{eq:detailed_dynamic_model}
\begin{aligned}
    & x_{t} = v_{t-1}\cos{(\phi_{t-1})}\Delta t + x_{t-1}, \\
    & y_{t} = v_{t-1}\sin{(\phi_{t-1})}\Delta t + y_{t-1}, \\
    & \phi_{t} = v_{t-1}\frac{\tan{(\delta_{t-1})}}{L}\Delta t + \phi_{t-1}, \\
    & v_{t} = a_{t-1}\Delta t + v_{t-1}, \\
\end{aligned}
\end{equation}
where $L$ denotes the vehicle wheelbase length.
The output vehicle waypoints and states $(x_t, y_t, \phi_t, v_t)$ are also passed through an MLP layer.
Then the resulted embedding $\mathbf{S}_t$ are concatenated with aforementioned $\mathbf{A}_{\mathcal{I}}$ yielding the LSTM input $\mathbf{i}$ for the next LSTM iteration.

\subsubsection{Differentiable Rasterizer}
\begin{figure}
\begin{minipage}{0.15\textwidth}
 \centering
 \subfloat[Vehicle shape\label{figure:agent_box_shape}]{\includegraphics[width=1\linewidth]{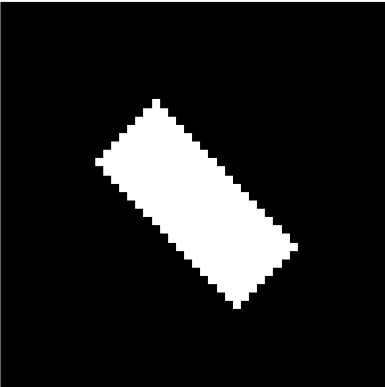}}
 \end{minipage} \hfill
\begin{minipage}{0.15\textwidth}
 \centering
 \subfloat[Rasterized shape\label{figure:agent_box_rasterized_shape}]{\includegraphics[width=1\linewidth]{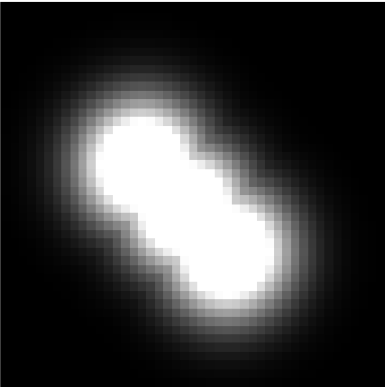}}
 \end{minipage} \hfill
\begin{minipage}{0.15\textwidth}
 \centering
 \subfloat[Gaussian kernels\label{figure:agent_box_rasterisation_comparison}]{\includegraphics[width=1\linewidth]{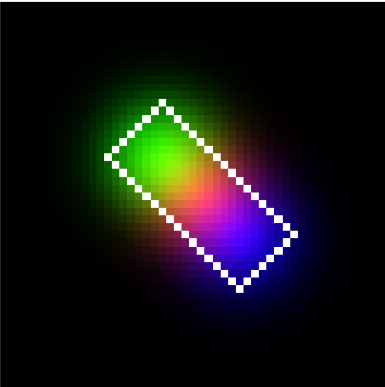}}
 \end{minipage} \hfill
 \caption{\small Visualization of rasterized images: (a) vehicle shape that we aim to rasterize; (b) the rasterized images by our differentiable rasterizer; (c) visualization of the Gaussian kernels used by the rasterizer in different colors.}
 \label{figure:rasterizer_visualization}
 \vspace{-0.4cm}
\end{figure}

During training time, the output trajectory points are rasterized into $N$ images.
As shown in Figure~\ref{figure:rasterizer_visualization}, unlike~\cite{cui2020ellipse}, we represent our ego vehicle using three 2D Gaussian kernels instead of one, which better sketch its shape.
Given the vehicle waypoints $\mathbf{s}_t = [x_t, y_t, \phi_t, v_t]^T$, vehicle length $l$, and width $w$, our rasterization function $\mathcal{G}$ rasterizes images as:
\begin{equation}
\label{eq:rasterisation_equations}
\mathcal{G}_{i,j}(\mathbf{s}_t) = \max_{k = 1,2,3}(\mathcal{N}(\mathbf{\mu}^{(k)}, \mathbf{\Sigma}^{(k)})),
\end{equation}
where each cell $(i, j)$ in $\mathcal{G}$ is denoted as $\mathcal{G}_{i,j}$.
Let $\mathbf{x}^{(k)}_t = [x_t^{(k)}, y_t^{(k)}]^T$ denote the center of the $k$-th Gaussian kernel of our vehicle representation. Then,
\begin{equation}
\begin{aligned}
\mathbf{\mu}^{(k)} &= \frac{1}{\rho}(\mathbf{x}^{(k)}_t - \mathbf{x}_0) + \mathbf{p}_0, \\
\mathbf{\Sigma}^{(k)} &= \mathbf{R}(\phi_t)^T\mathrm{diag}(\sigma_l, \sigma_w)\mathbf{R}(\phi_t),
\end{aligned}
\end{equation}
where $(\sigma_l, \sigma_w) = (\frac{1}{3}\alpha l, \alpha w)$,  $\alpha$ is a fixed positive scaling factor, and $\mathbf{R}(\phi)$ represents a rotation matrix constructed from the heading $\phi$.
Note that the Gaussian distributions are not truncated by the vehicle box-like shape allowing the model to learn how human drivers keep a safe distance to obstacles or road boundaries.
Our experiments also demonstrate that this design choice helps us achieve better performance, as shown in Section~\ref{subsubsec:rasterizer_design_analysis}.

\subsubsection{Loss}
Benefiting from the LSTM decoder and differentiable rasterizer, our trajectory imitation loss is simple and can be expressed in analytic form as:
\begin{equation}
\label{eq:imitation_loss}
\mathcal{L}_\mathrm{imit}= \sum_{t=0}^{N - 1} \mathbf{\lambda}\norm{\mathbf{s}_t - \hat{\mathbf{s}}_t}_2,
\vspace{-0.15cm}
\end{equation}
where $\norm{\cdot}_2$ is the L2 norm, $\mathbf{\lambda}$ are the weights, and $\hat{\mathbf{s}}_t$ denotes the corresponding ground truth vehicle waypoints.

\begin{figure}
\begin{minipage}{0.232\textwidth}
 \flushright
 \subfloat[Obstacle mask
 \label{figure:obstacle_loss_mask}]{\includegraphics[width=0.8\linewidth]{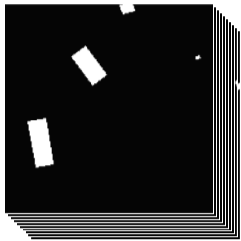}}
\end{minipage}\hfill
\begin{minipage}{0.232\textwidth}
 \flushleft
 \subfloat[Route mask
 \label{figure:routing_loss_mask}]{\includegraphics[width=0.8\linewidth]{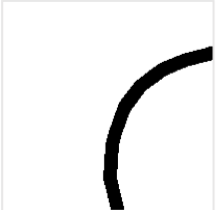}}
 \end{minipage}
\begin{minipage}{0.232\textwidth}
 \flushright
 \subfloat[Road mask \label{figure:off_road_loss_mask}]{\includegraphics[width=0.8\linewidth]{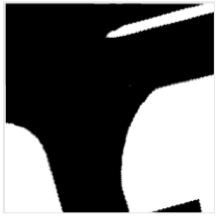}}
\end{minipage}\hfill
\begin{minipage}{0.232\textwidth}
 \flushleft
 \subfloat[Traffic signal mask \label{figure:traffic_light_loss_mask }]{\includegraphics[width=0.8\linewidth]{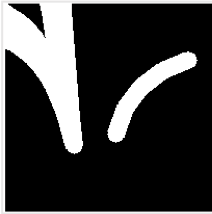}}
 \end{minipage}
 \caption{\small Visualization of the binary masks we used in our task losses: (a) the obstacle masks that push our vehicle away from obstacles; (b) The route masks that keep our vehicle on the planned route by penalizing the behaviors that cause any portion of our vehicle that does not overlap with the planned route; (c) The road masks that keep our vehicle within drivable areas; (d) The masks that make our vehicle stop in front of traffic signals when they are red.}
 \label{figure:task_loss_masks_representation}
 \vspace{-0.4cm}
\end{figure}

Besides the imitation loss, four task losses are introduced to prevent our vehicle from undesirable behaviors, such as obstacle collision, off-route, off-road, or traffic signal violation. $\mathcal{T}^\mathrm{obs}$, $\mathcal{T}^\mathrm{route}$, $\mathcal{T}^\mathrm{road}$, and $\mathcal{T}^\mathrm{signal}$ are corresponding binary masks, as shown in Figure~\ref{figure:task_loss_masks_representation}.
We assign ones to all the cells that our vehicle should avoid in our binary masks.
The following losses are included:
\begin{equation}
    \begin{split}
    \mathcal{L}_\mathrm{obs} = &\sum_{t=0}^{N-1}\frac{1}{WH}\sum_{i}\sum_{j} \mathcal{G}_{i,j}\mathcal{T}^\mathrm{obs}_{i,j}, \\
    \mathcal{L}_\mathrm{route} = &\sum_{t=0}^{N-1}\frac{1}{WH}\sum_{i}\sum_{j} \mathcal{G}_{i,j}\mathcal{T}^\mathrm{route}_{i,j}, \\
    \mathcal{L}_\mathrm{road} = &\sum_{t=0}^{N-1}\frac{1}{WH}\sum_{i}\sum_{j}\mathcal{G}_{i,j}\mathcal{T}^\mathrm{road}_{i,j}, \\
    \mathcal{L}_\mathrm{signal} = &\sum_{t=0}^{N-1}\frac{1}{WH}\sum_{i}\sum_{j}\mathcal{G}_{i,j}\mathcal{T}^\mathrm{signal}_{i,j}. \\
    \end{split}
\end{equation}

Overall, our total loss is given by:
\begin{equation}
    \mathcal{L} = 
    \mathcal{L}_{imit} + \lambda_\mathrm{task}(\mathcal{L}_\mathrm{obs} + \mathcal{L}_\mathrm{route} +\mathcal{L}_\mathrm{road} + \mathcal{L}_\mathrm{signal}),
\end{equation}
where $\lambda_\mathrm{task}$ is an empirical parameter.
Like~\cite{bansal2019chauffeurnet}, we randomly drop out the imitation losses for some training samples to further benefit from the task losses.

\subsection{Data Augmentation}
To address the aforementioned distributional shift issue, we further introduce an iterative feedback synthesizer to enrich the training dataset besides the random horizontal perturbations~\cite{bansal2019chauffeurnet}.
Both data augmentation methods help our agent broaden its experience beyond the demonstrations.

\subsubsection{Random Synthesizer}
Similar to \cite{bansal2019chauffeurnet}, our synthesizer perturbs trajectories randomly, creating off-road and collision scenarios.
The start and end points of a trajectory are anchored, while we perturb one of the points on the trajectory and smooth across all other points.
Only realistic perturbed trajectories are kept by thresholding on maximum curvature.

\subsubsection{Feedback Synthesizer}
\begin{figure}
	\centering
	\includegraphics[width=\linewidth]{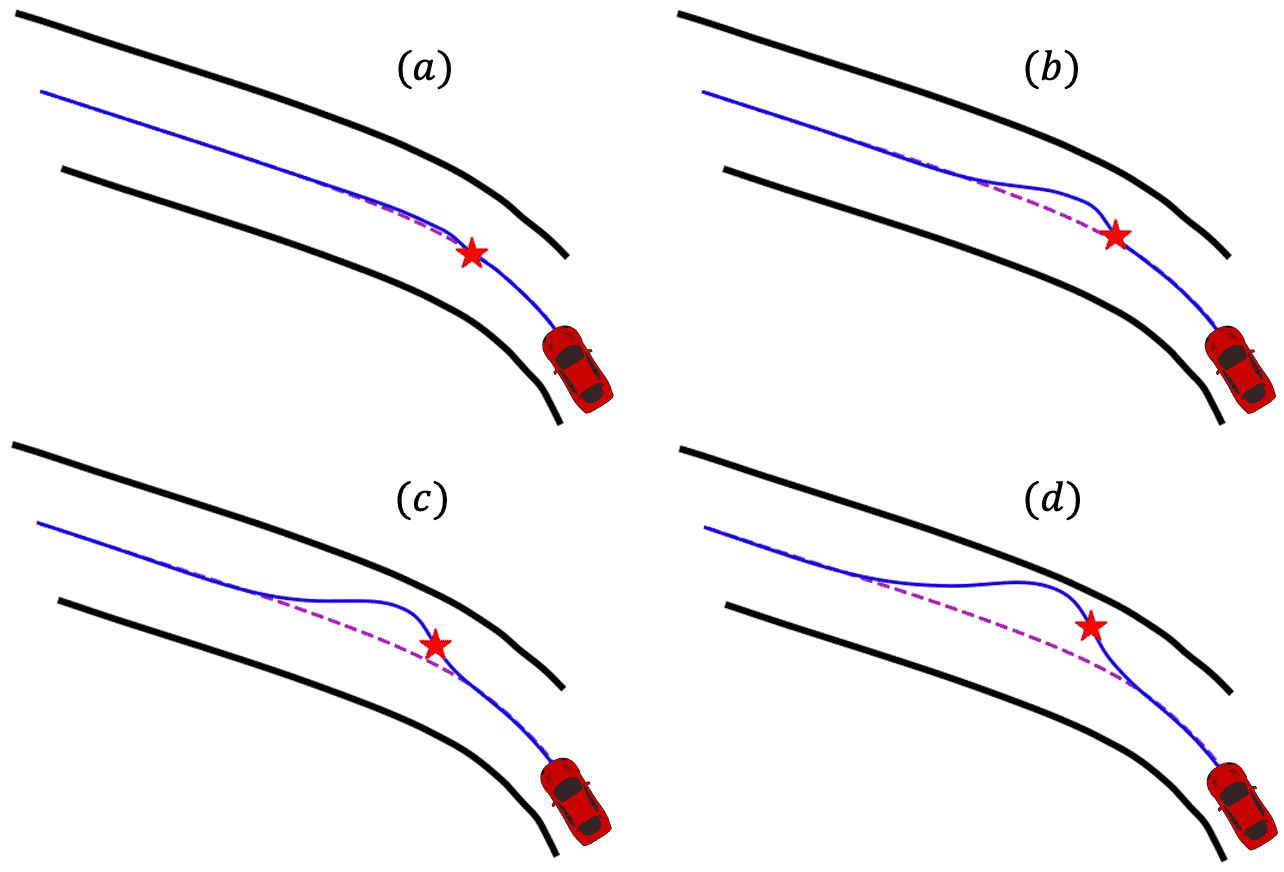}
	\caption{\small The illustration of our feedback synthesizer. The ``star'' marks where we construct a new state by sampling an action using the current policy. We fit a smooth trajectory with necessary kinematic constraints after the ``star'' point, therefore avoiding any expert agents' dependence. (a) - (d) demonstrate the trajectories generated when $T=2, 4, 6, 8$.
	}
	\label{fig:feedback}
	\vspace{-0.4cm}
\end{figure}

The random synthesizer above achieves impressive performance improvement, as we shall see in Section \ref{section:exp}.
However, it fails to address one important issue due to its random nature, the bias in probability distributions of the synthesized states.
To this end, following the principle of DAgger \cite{ross2011reduction}, we propose an iterative feedback synthesizer that generates the states that our agent is more likely to encounter by sampling the data based on the current policy.
Thus, we seek to improve our policy by adding these states and their corresponding actions into the training dataset.
If we denote our policy model at $i$-th iteration as $\pi_i$, given any start point $t_0$ in a trajectory, we can sample future $T$ steps using $\pi_i$.
This constructs a few new states that our agent is likely to encounter.
To demonstrate how our agent should react to these states that are deviated from our optimal behavior, we propose to fit a smooth trajectory with necessary kinematic constraints.
Figure \ref{fig:feedback} illustrates the deviated trajectories given different $T$ values from 2 to 8. A complete step-by-step overview is detailed in Algorithm \ref{alg:iter_feedback}.
The whole process is fully automatic and doesn't rely on the participation of any expert agent.
Note that some synthesized samples may be discarded when the solver fails to compute feasible and smooth trajectories.

\begin{algorithm}
	\caption{Feedback Synthesizer based Iterative Training}
	\label{alg:iter_feedback}
	\begin{algorithmic}[1]
		\State Input $\mathcal{D} = \mathcal{D}_0$. \Comment {trajectories from human drivers}
		\State Train policy $\pi_0$ on $\mathcal{D}$.
		\For{$i=1$ to $K$}
		\State Sample $T$-step trajectories using $\pi_{i-1}$ from $\mathcal{D}_0$.
		\State Get dataset $D_i = \{(s, a)\}$ of visited states by $\pi_{i-1}$ and actions generated by smoothing.
		\State Aggregate datasets $\mathcal{D} \leftarrow \mathcal{D} \cup \mathcal{D}_i$
		\State Train policy $\pi_i$ on $\mathcal{D}$.
		\EndFor
	\end{algorithmic}
\end{algorithm}
\subsection{Post-processing Planner}

This section introduces our post-processing planner, a new architecture that aims to offer better comfort and improve driving safety as a gatekeeper.
The implementation has been released in Apollo 6.0 at \url{https://github.com/ApolloAuto/apollo/tree/master/modules/planning}.

\begin{figure}
\centering
\begin{minipage}{0.23\textwidth}
 \centering
 \captionsetup{justification=centering}
 \subfloat[Planner architecture\label{figure:hybrid_structure}]{\includegraphics[width=\linewidth]{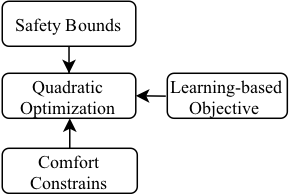}}
 \end{minipage}
\begin{minipage}{0.23\textwidth}
 \centering
 \captionsetup{justification=centering}
 \subfloat[Bounds and  objectives\label{figure:hybrid_path}]{\includegraphics[width=\linewidth]{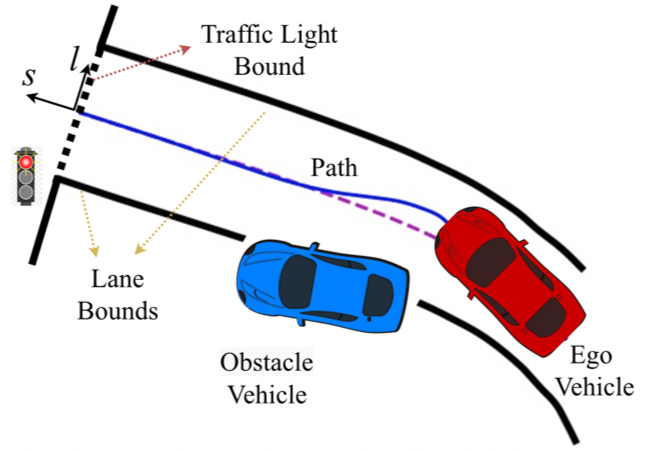}}
 \end{minipage}
 \caption{\small The illustration of our post-processing planner design. (a) We formulate a joint optimization problem by incorporating safety and comfort bounds into the framework while taking the learning-based planner's trajectory output as the optimization objective. (b) An example illustrates how we build safety and comfort bounds and the optimization objective.}
 \vspace{-0.4cm}
\end{figure}

The traditional path planning task is typically formulated as a quadratic optimization problem where lane boundaries, collision, traffic lights, and other driving conditions are formulated as bounding constraints \cite{zhang2020optimal}.
Our new architecture is a joint optimization framework based on cues from both the kinematics and dynamics constraints and the above learning-based planner's output waypoints, as shown in Figure \ref{figure:hybrid_structure}.
In the optimization framework, the bounding constraints remain unchanged.
Meanwhile, the output waypoints of the learning-based planner are taken as its optimization objective.
Finally, the joint optimization framework produces eventual path trajectories that are fed to a control module.

Figure \ref{figure:hybrid_path} illustrates an example on a real road.
Our vehicle is approaching an intersection that generates a bound in front because of traffic signals.
Lane boundaries and dynamic obstacles on the road also compose safety bounds in our optimization problem.
Smoothness constraints ensure driving comfort.
As opposed to the centerline (the default optimization objective in a traditional planner), the learning-based planner's waypoints guide our ego vehicle to keep a safe distance from the obstacle and successfully overtake it, which is similar to human driving behaviors.
The final path, generated by the post-processing planner, follows our learning-based planner's high-level decision to overtake the obstacle vehicle while ensuring safety and comfort.

\begin{table*}
\centering
\caption{\small Comparisons of different model configurations. The baseline model (M0) with pure imitation learning losses only achieves 15.7\% ADS pass rate, while our proposed learning-based planner (M2) achieves 70.0\% ADS pass rate. We further improve our ADS pass rate and comfort score by having an optional optimization-based post-processing planner (M3).}
\begin{tabular}{p{0.04\linewidth} | p{0.525\linewidth} | p{0.1\linewidth} | p{0.085\linewidth} | p{0.05\linewidth} | p{0.05\linewidth}}
 \hline
 Model & Description & Training Loss & Datasets ($D_o, D_r, D_f$) & Pass Rate & Comfort Score\\
 \hhline{=:=:=:=:=:=}
  $\mathbf{M0}$ & \text{CNN backbone + LSTM decoder + imitation loss} & $\mathcal{L}_{imit} $ & \checkmark, $\times$, $\times$ & 15.71\% & 0.1048 \\
 \hhline{-|-|-|-|-|-}
  $M0a$ & \text{M0 + differentiable rasterizer and task losses + random synthesizer} & $\mathcal{L}_{imit}  + \mathcal{L}_{task}$ & \checkmark, \checkmark, $\times$ & 54.28\% & 0.0228 \\
 \hhline{-|-|-|-|-|-}
  $M0b$ & \text{M0 + spatial attention branch} & $\mathcal{L}_{imit} $ & \checkmark, $\times$, $\times$ & 10.00\% & 0.1105 \\
 \hhline{=:=:=:=:=:=}
  $\mathbf{M1}$ & $M0a + M0b$ & $\mathcal{L}_{imit}  + \mathcal{L}_{task}$ & \checkmark, \checkmark, $\times$ & 55.71\% & 0.0214 \\
 \hhline{-|-|-|-|-|-}
  $M1a$ & \text{M1 with attention applied directly on raw image rather than intermediates $\mathbf{F}_{\mathcal{I}}$} & $\mathcal{L}_{imit}  + \mathcal{L}_{task}$ & \checkmark, \checkmark, $\times$ & 42.85\% & 0.0070 \\
  \hhline{-|-|-|-|-|-}
  $M1b$ & \text{M1 with attention module implemented inside the backbone, not as a branch} & $\mathcal{L}_{imit}  + \mathcal{L}_{task}$ & \checkmark, \checkmark, $\times$ & 51.43\% & 0.0103 \\
 \hhline{-|-|-|-|-|-}
  $M1c$ & \text{M1 with truncated rasterization by vehicle shapes} & $\mathcal{L}_{imit}  + \mathcal{L}_{task}$ & \checkmark, \checkmark, $\times$ & 48.57\% & 0.0201 \\
 \hhline{=:=:=:=:=:=}
  $\mathbf{M2}$ & \text{M1 + feedback synthesizer} & $\mathcal{L}_{imit}  + \mathcal{L}_{task}$ & \checkmark, \checkmark, \checkmark & 70.00\% & 0.0150 \\
 \hhline{=:=:=:=:=:=}
  $\mathbf{M3}$ & \text{M2 + post-processing planner} & / & / & 94.29\% & 0.1333 \\
 \hline
\end{tabular}
\vspace{-0.4cm}
\label{table:ablation_tests}
\end{table*}

\section{experiments}
\label{section:exp}

\subsection{Implementation Details}

We have a squared BEV image with $W\!=\!200$, $H\!=\!200$, and $\rho\!=\!0.2 m/\mathrm{pixel}$.
In the image, our ego-vehicle is placed at $i_0\!=\!100$, $j_0\!=160$ in image coordinates.
For the training data, we use a perception system in Apollo \cite{baidu-apollo} that processes raw sensor measurements and produces object detection, tracking, and prediction results.
Two seconds of history or predicted trajectories are rasterized in the corresponding image channels.
Our learning-based planning module outputs trajectories with a time horizon as $N = 10$ and $\Delta t = 0.2s$.
Our models were trained using the Adam optimizer with an initial learning rate of 0.0003.

 

\subsection{Dataset and Augmented Data}
Unlike prior works \cite{codevilla2018end,codevilla2019exploring,prakash2020exploring} that exploit a driving simulator and collect simulated driving data, we recruited our self-driving test vehicles and collected a total of 400 hours' driving data demonstrated by human-drivers in southern San Francisco bay area.



After necessary data pre-processing, we keep about 250k frames as our original training data before any data augmentation, denoted as $D_o$.
Then, additional 400k frames are generated by our random synthesizer, denoted as $D_r$. When the feedback steps $T=5$, our feedback synthesizer generates about 465k frames in one iteration, and they are denoted as $D_f$.
To keep the training procedure efficient, we let the number of iterations $K\!=\!1$ in the feedback synthesizer.
Table~\ref{table:ablation_tests} lists different configurations of training data when we evaluate different methods.

\begin{figure}
  \includegraphics[width=0.95\columnwidth]{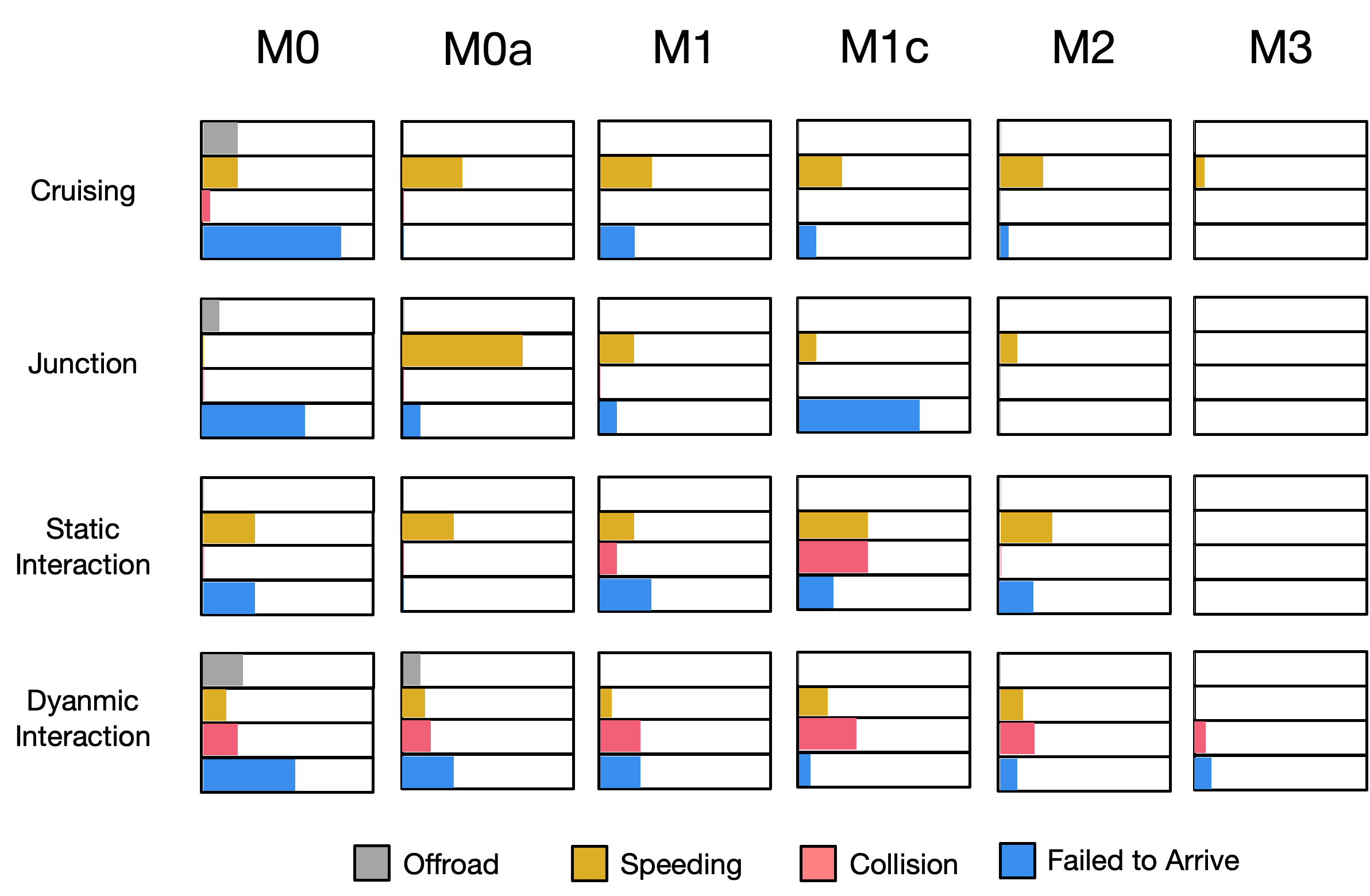}
  \caption{\small Detailed analysis of failure reasons of different model configurations under different driving scenarios.}
  \label{fig:closed_loop_results}
  \vspace{-0.4cm}
\end{figure}

\subsection{Evaluation Scenarios}
\label{subsec:evaluation_scenarios}
To fully evaluate the proposed methods, we use the Apollo Dreamland simulator \cite{baidu-apollo} and carefully created 70 ADS.
Each ADS lasts about 30-90 seconds and includes one or more driving scenarios. The ADS are either handcrafted by our scene designers or taken from logs from real driving data (separate from the training data).

The driving scenarios in the 70 ADS can be roughly categorized into four types:
\begin{itemize}
\item \textbf{Cruising}: normal cruising in straight or curved roads without other traffic participants.
\item \textbf{Junction}: junction related scenarios including left or right turns, U-turns, stop before a traffic signal, etc. 
\item \textbf{Static Interaction}: interaction with static obstacles, such as overtaking a stopped car.
\item \textbf{Dynamic Interaction}: interaction with dynamic obstacles, for example, overtaking a slow vehicle.
\end{itemize}
They constitute a comprehensive driving test challenge specifically designed to evaluate the fundamental driving capabilities of a self-driving vehicle.
Using these test scenarios, we carry out closed-loop evaluations where a control module executes our output waypoints, and the results are summarized in Table~\ref{table:ablation_tests}.



\subsection{Evaluation Metrics}
Besides the pass or the success rate that prior works \cite{bansal2019chauffeurnet,codevilla2019exploring} focus on,
we propose to evaluate how comfortable the driving is and formulate a new comfort score.
We assume that human drivers deliver a comfortable driving experience.
The comfort score is calculated based on how similar our driving states are to human drivers. Particularly, we record the probabilities of a set of angular velocity ($\mathrm{rad}/s$) and jerk ($m/s^3$) values $(\omega, j)$ in our human-driving data $D_o$.
Given $(\omega, j)$ from certain driving data of an agent, we define the comfort score $c$ as:
\begin{equation}
c = \frac{\sum_{i=1}^{n}P(\omega ,j|D_o)}{n},
\end{equation}
where $P(\omega ,j|D_o)$ is the probabilities of values $(\omega, j)$ happened in human-driving data $D_o$. $n$ is the number of frames, and $(\omega, j)$ are the corresponding values in our test driving data.
Given $(\omega, j)$, when we look up the corresponding probabilities, $\omega$ and $j$ are discretized into bins with sizes of 0.1 and 1.0, respectively.

Our ADS pass rate is based on a set of safety-related metrics, including collision, off-road, speeding, traffic-light violation, etc. When any of them is violated, we grade the entire ADS as failed. Of course, the agent has to reach the scheduled destination in a limited time to be considered successful, too.





\subsection{Runtime Analysis}
\label{subsec:run_time}
We evaluated our method's runtime performance with an Nvidia Titan-V GPU, Intel Core i7-9700K CPU, and 16GB Memory.
The online inference time per frame is 10ms in rendering, 22 ms in model inference, and 15 ms (optional) in the post-processing planner.
Note that our model inference time is much shorter than the prior work \cite{bansal2019chauffeurnet}.

\begin{figure}
\begin{minipage}{0.15\textwidth}
\centering
\subfloat[M0b\label{figure:M0b_attention_heatmap}]{\includegraphics[width=1\linewidth]{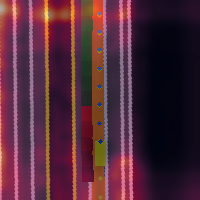}}
\end{minipage}\hfill
\begin{minipage}{0.15\textwidth}
\centering
\subfloat[M1a\label{figure:M1a_attention_heatmap}]{\includegraphics[width=1\linewidth]{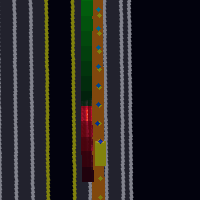}}
\end{minipage}\hfill
\begin{minipage}{0.15\textwidth}
\centering
\subfloat[M1b\label{figure:M1b_attention_heatmap}]{\includegraphics[width=1\linewidth]{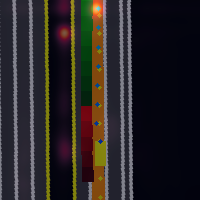}}
\end{minipage}\hfill
\caption{\small Comparisons of the attention heatmaps generated by different models in a dynamic interaction scenario. The corresponding heatmap from M1 can be found in Figure~\ref{figure:attention}(a). Output and ground truth trajectories are illustrated as blue and yellow dots, respectively.}
\label{figure:attention_heatmap_comparison}
\vspace{-0.4cm}
\end{figure}

\begin{figure}
	\centering
	\includegraphics[width=\linewidth]{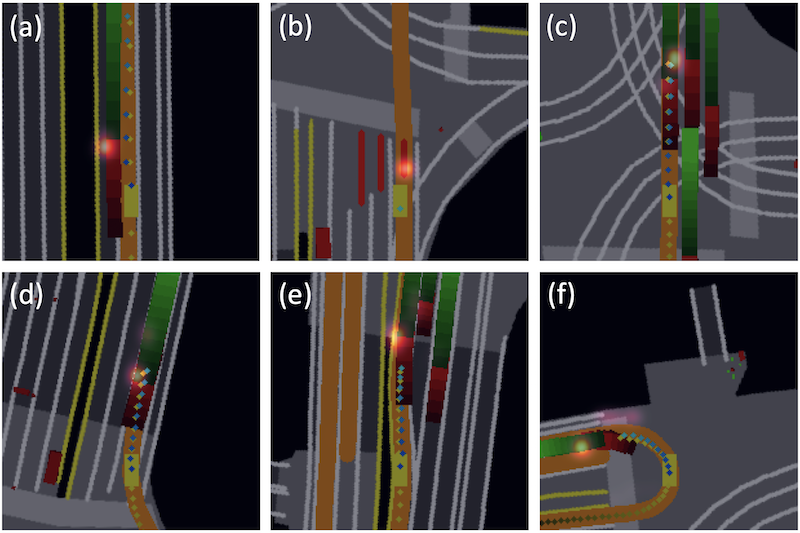}
 \caption{\small Visualization of the attention heatmaps from M1 in different scenarios: (a) dynamic nudge; (b) stopping at a red light; (c) following a leading vehicle; (d) right turn; (e) yielding to a vehicle; (f) U-turn.}
 \label{figure:attention}
 \vspace{-0.4cm}
\end{figure}

\subsection{Comparison and Analysis}
To demonstrate each of the above contributions' effectiveness, we show the ADS pass rate and comfort score with different methods in Table~\ref{table:ablation_tests}.
Also, the proportion of scenarios where we failed due to any violation of each particular rule in different driving scenarios using different methods are shown in Figure~\ref{fig:closed_loop_results}.
Detailed analysis is discussed as follows.

\subsubsection{Task Losses}
\label{subsubsec:task_loss_effect_analysis}
By comparing M0a with M0, it is observed that the numbers of scenarios where we failed due to off-road or collision reasons decrease.
A similar trend can be found by observing the pass rate from M0b and M1.
In particular, we add the differentiable rasterizer and task losses in our models M0a and M1.
They play an important role in keeping us within drivable areas and avoiding obstacles.

\subsubsection{Spatial Attention}
\label{subsubsec:attention_ads_pass_rate_analysis}
The spatial attention in M0b gives a slightly worse pass rate with pure imitation losses comparing to M0.
It implies that the attention can not improve the performance by itself without the important task losses.
It makes sense since the task losses involve obstacles, traffic signals, and other traffic participants in explicit forms.
They help the network locate the objects that it should pay attention to.
In Figure~\ref{figure:M0b_attention_heatmap}, we also find the attention heatmaps failed to converge in M0b.
By comparing M1 with M0a, we find slight improvements in pass rate by having the spatial attention module.
Also, we visualize the attention heatmaps from M1 in Figure~\ref{figure:attention}, and they are easy to interpret and discover the attentive objects.
To further verify our attention module design, we did experiments with M1a and M1b comparing with M1.
M1a applies attention weights directly on raw image pixels resulting in higher computation complexity.
In our experiment, M1a concluded with a lower ADS pass rate and less interpretable attention heatmaps, as shown in Figure~\ref{figure:M1a_attention_heatmap}.
M1b feeds the attention features back to the CNN backbone, also resulting in worse ADS pass rate and wrongly focused attention heatmaps as shown in Figure~\ref{figure:M1b_attention_heatmap}.
Our design's inspiration is that keeping original CNN features in a separate branch is good for providing more meaningful features for downstream modules, therefore better for the overall performance.

\subsubsection{Differentiable Rasterizer}
\label{subsubsec:rasterizer_design_analysis}
When rasterizing vehicle boxes, it's intuitive to truncate the Gaussian distributions by box boundaries ensuring a precise representation.
However, while we test it in experiment M1c, the experimental results suggest otherwise.
M1c's ADS pass rate drops slightly compared to M1, the model without the truncation step.
More importantly, M1 has fewer failed scenarios, especially in the collision category, as shown in Figure~\ref{fig:closed_loop_results}.
It implies the long tail in distributions helps with obstacle avoidance. Besides this, we also find that M1c has a higher failure rate in junction scenarios where our vehicles stopped before traffic lights but failed to restart when they turned green.
A possible reason is that the overlap between the long tail in distributions and the traffic signal mask provides more hints to the model and helps it learn the stop's causality.


\subsubsection{Feedback Synthesizer}
As we mentioned, our feedback synthesizer addresses the fundamental issue with imitation learning, the distributional shift.
As a result, M2 improved the overall performance comparing to M1.
Specifically, M2 achieves 70.0\% ADS pass rate, about a 25\% improvement over M1.
The number of failure ADS decreases across almost all different scenarios or failure reasons, as shown in Figure~\ref{fig:closed_loop_results}.

\subsubsection{Post-processing Planner}
Our post-processing planner (M3) is designed as a gatekeeper with safety constraints.
It's not surprising that it achieves a higher ADS pass rate, 94.29\%.
Better yet, the post-processing planner also significantly improves the comfort score with its smoothness constraints in the joint optimization framework.
Note that we find pure imitation learning methods (M0 or M0b) without any data augmentation achieve similar comfort scores to M3.
It seems that data augmentation hurts driving comfort, and our newly proposed comfort metric reveals this. 
We also evaluated the Post-processing Planner without the trajectory output from learning-based planner but using lane center-line as guidance (Traditional Planner). It has 88.57\% pass rate and 0.1342 comfort score. Compared with the Post-processing Planner, the Traditional Planner has lower pass rate, as we expect that learning-based planner’s high-level decision does perform better in  complex scenarios.

\subsubsection{Failure Modes}
From Figure~\ref{fig:closed_loop_results}, it is easy to tell that speeding issues have already become a major problem for our learning-based planner, M2.
These results seem to suggest that further improvements may be possible by adding a speeding loss.
Another important failure reason is non-arrival.
It happens when our ego vehicle gets stuck somewhere or has a too low speed.
It probably means that we could extend our work and encourage our vehicle to keep moving forward at a steady speed by having a new loss which could be similar to the reward in reinforcement learning.

For our post-processing planner M3, safety constraints were implemented. We examined two failure scenarios that are related to collision issues.
Our vehicle is hit by a car behind it in both scenarios, which is a rear-end collision.
Also note that, due to the limitation of our test environment, all other vehicles in the ADS only move forward by following scheduled routes and are not intelligent to avoid our ego-vehicle.
This increases the chance of rear-end collisions, especially when our vehicle is slower than expected.



\section{conclusion and Future Work}
We have presented an imitation learning based planner designed for autonomous driving applications.
We demonstrate that the overall ADS pass rate can be improved using necessary data augmentation techniques, including a novel feedback data synthesizer.
Benefiting from the differentiable rasterizer, our learning-based planner runs inference in real-time, yielding output either directly used by a downstream control module or further fed into an optional post-processing planner.
Task losses and spatial attention are introduced to help our system reason critical traffic participants on the road.
Therefore, the full exploitation of the above modules enables our system to avoid collisions or off-road, smoothly drive through intersections with traffic signals, and even overtake slow vehicles by dynamically interacting with them.
In the future, we plan to collect more real-road driving data, further strengthening our imitation learning capabilities.
Also, the introduction of reinforcement learning maybe another direction to improve the overall performance to a higher level.

\begin{small}
\section*{ACKNOWLEDGMENT}
We would like to thank our colleagues for their kind help and support throughout the project. Yiqun Fu helped with data pre-processing. Leonard Liccardo,  Wesley Reynolds, Christopher Heineman, and Douglas Wells helped with training data collection.
\end{small}

\bibliographystyle{IEEEtran}
\bibliography{IEEEabrv,sections/refs}

\end{document}